\documentclass[conference]{IEEEtran}
\IEEEoverridecommandlockouts
\usepackage{cite}
\usepackage{amsmath,amssymb,amsfonts}
\usepackage{algorithm}
\usepackage{algpseudocode}
\usepackage{graphicx}
\usepackage{changes}
\usepackage{textcomp}
\usepackage{xcolor}
\usepackage{subcaption}
\usepackage{mymacros}
\usepackage{comment}

\usepackage{geometry}
\begin{document}
\title{\vspace{0.63cm}Enforcing safety for vision-based controllers via Control Barrier Functions and Neural Radiance Fields
\thanks{C. Dawson is supported by the NSF GRFP under Grant No. 1745302. The Defense Science and Technology Agency in Singapore and MIT-IBM provided funds to assist the authors with their research, but this article solely reflects the opinions and conclusions of its authors and not DSTA Singapore, the Singapore Government, or MIT-IBM.}
}

\author{\IEEEauthorblockN{Mukun Tong}
\IEEEauthorblockA{\textit{Dept. of Automation} \\
\textit{Tsinghua University}\\
Beijing, China \\
\texttt{tmk19@mails.tsinghua.edu.cn}}
\and
\IEEEauthorblockN{Charles Dawson}
\IEEEauthorblockA{\textit{Dept. of Aeronautics and Astronautics} \\
\textit{MIT}\\
Cambridge, USA \\
\texttt{cbd@mit.edu}}
\and
\IEEEauthorblockN{Chuchu Fan}
\IEEEauthorblockA{\textit{Dept. of Aeronautics and Astronautics} \\
\textit{MIT}\\
Cambridge, USA \\
\texttt{chuchu@mit.edu}}
}

\maketitle

\begin{abstract}
To navigate complex environments, robots must increasingly use high-dimensional visual feedback (e.g. images) for control. However, relying on high-dimensional image data to make control decisions raises important questions; particularly, how might we prove the safety of a visual-feedback controller? Control barrier functions (CBFs) are powerful tools for certifying the safety of feedback controllers in the state-feedback setting, but CBFs have traditionally been poorly-suited to visual feedback control due to the need to predict future observations in order to evaluate the barrier function. In this work, we solve this issue by leveraging recent advances in neural radiance fields (NeRFs), which learn implicit representations of 3D scenes and can render images from previously-unseen camera perspectives, to provide single-step visual foresight for a CBF-based controller, where the CBFs possess a discrete-time nature. This novel combination is able to filter out unsafe actions and intervene to preserve safety. We demonstrate the effect of our controller in real-time simulation experiments where it successfully prevents the robot from taking dangerous actions.
\end{abstract}


\section{Introduction \& Related Work}

Robots are increasingly making use of visual feedback to explore novel, complex environments, particularly in GPS-denied environments, and recent works have demonstrated the potential for using this visual feedback directly for control~\cite{pmlr-v164-margolis22a,malis2002survey}. However, before we can use visual data to make control decisions in safety-critical robotic systems, we must be able to certify the safety of vision-based controllers. For instance, imitation learning-based pixel controllers have been a popular way of combining control and vision \cite{pan2017agile}, but lack safety guarantees,  which can bring risks to the system.

Unfortunately, ensuring safety for visual-feedback controllers is a challenging problem, since this issue lies in the intersection of control theory and computer vision. To solve this problem, we bring together two powerful concepts from each of these disciplines. Control barrier functions (CBFs) are powerful tools from control theory for proving the safety of a feedback controller~\cite{ames_cbf, dawson2022safe}; in contrast with reachability-based approaches to safety verification~\cite{katz_image_based}, CBFs require us to consider only a single-step horizon, reducing computation time and allowing them to be used for real-time control. Unfortunately, CBFs typically require a known model for how control actions affect future measurements of the CBF. For systems with state feedback, such a model is often available, and approximate models are available for some types of high-dimensional observations such as Lidar~\cite{dawson22perception}, but the lack of a predictive model for visual feedback has so far prevented the use of CBFs in the visual-feedback setting. To close this gap, we turn to neural radiance fields (NeRFs), a recently developed technique from the computer vision community where a neural network is used to learn an implicit representation of a 3D scene~\cite{mildenhall2020nerf}. NeRFs can be used to re-render a scene from a previously-unseen perspective, allowing a robot to predict the image it would observe if it were to move from its current viewpoint. Recent work on NeRF-enabled SLAM~\cite{zhu2022nice,SucarICCV2021_imap} has shown that neural implicit scene representations can be trained online for novel environments, opening up a range of robotics applications for this technology.

Our key insight in this paper is that implicit neural scene representations (like NeRFs) can provide \textit{foresight} for vision-based controllers; they allow a robot to predict the effect of its actions on future image observations, allowing it to filter out actions that produce unsafe future observations. This insight leads us to the following contributions:
\begin{enumerate}
    \item We present a novel CBF-based controller that ensures safety for a controller operating solely on high-dimensional visual feedbacks.
    \item We combine discrete-time CBFs with NeRFs to enable single-step visual foresight, allowing our controller to determine whether an action is safe or unsafe using only a single-step horizon.
    \item We demonstrate that this controller can run in simulation at real-time rates (improving on previous NeRF-based works by several orders of magnitude).
\end{enumerate}

\section{Background \& Preliminaries}

In this section we will briefly introduce NeRFs and CBFs, upon which our visual-foresight CBF controller builds.

\subsection{Neural Radiance Fields} 

Neural Radiance Fields (NeRFs) are a recently-developed method to synthesize views of 3D scenes by learning a latent, continuous representation of a 3D volume. NeRFs represent a scene with a fully connected deep network that predicts the density and color of points in 3D space as a function of location and viewing direction. A NeRF is typically trained via a reconstruction loss on a series of example images, and the scene can be re-rendered from a new perspective using differentiable ray-tracing~\cite{mildenhall2020nerf}.

Recent works such as iMAP~\cite{SucarICCV2021_imap} and NICE-SLAM~\cite{zhu2022nice} have applied NeRFs to SLAM (Simultaneous Localization And Mapping) problems in robotics. By relying on a pre-trained network to warm-start the learning process, NeRFs can be fine-tuned online to represent novel environments, while simultaneously optimizing the parameters of the scene representation and the estimated camera pose allows a robot to simultaneously estimate its own position. Other works have attempted to use NeRFs as the basis for a trajectory optimization pipeline~\cite{Adamkiewicz2022nerf_navigation}, but since trajectory optimization over a long horizon requires hundreds of calls to the underlying NeRF, these techniques are extremely computationally expensive.

\subsection{Control Barrier Functions}

Safety is a recurring problem in robotics: given a potentially nonlinear dynamical system, how can we choose control inputs to ensure that this system will remain safe? Control barrier functions (CBFs) have emerged to handle this problem analogously to how Control Lyapunov functions (CLFs) may be used to prove the stability of a control system~\cite{ames_cbf,dawson2022safe}. We will briefly review the theory of CBFs here; a more thorough survey can be found in~\cite{dawson2022safe}.

Consider a discrete-time dynamical system with state $x \in \mathcal{X} \subseteq \R^n$ and control $u \in \mathcal{U} \subseteq \R^m$, dynamics $x_{t+1} = f(x, u)$, and observations in the form of $w \times h$-size RGBd images of the robot's environment $y = o(x) \in \R^{w \times h \times 4}$. CBFs have been applied previously to Lidar feedback~\cite{dawson22perception}, but we are not aware of any other work considering the case of visual feedback without assuming access to a bounded-error state-estimation oracle~\cite{pmlr-v155-dean21a}. Let us denote as $\mathcal{X}_u \subset \mathcal{X}$ a set of \textit{unsafe} states that is disjoint from the set of admissible initial conditions $\mathcal{X}_0\subseteq\mathcal{X}\backslash\mathcal{X}_u$, and let us assume that we have some function $h$ that can distinguish between safe and unsafe states based on the observations $y$; i.e. $h(o(x)) > 0 \iff x \in \mathcal{X}_u$ and $h(o(x)) \leq 0$ otherwise. Further, let us assume that we have access to some \textit{nominal policy} $\pi_0(y)$. This nominal policy may be wildly unsafe, and so the goal of a CBF is to find some small control intervention that ensures the system remains safe.

The key result from CBF theory~\cite{Agrawal17dt} is that this safety-preserving control action can be found by solving the optimization problem:
\begin{subequations}
\begin{align}
    u = \argmin_u \quad & ||u - \pi_0(y_t)||^2 \label{cbf_optimization}\\
    \text{s.t.} \quad & h \circ o \circ f(x_t, u) \leq \alpha h \circ o(x)\label{cbf_constraint}
\end{align}
\end{subequations}
where $\circ$ denotes function composition and $\alpha \in (0, 1)$ is a rate-determining constant. If a feasible $u$ can always be found, then $h$ is a valid CBF and acts as a certificate to prove the safety of the control system. In this case, $h$ can be used to ``filter'' any potentially unsafe input, as may arise in a reinforcement learning or human teleoperation context.

Of course, since the constraint~\eqref{cbf_constraint} requires predicting the image $o\circ f(x_t, u)$ that would be observed if the robot were to take action $u$, CBFs have not previously seen much use in the direct visual-feedback setting. Some papers make promising steps towards understanding the case where visual feedback is used to for state estimation alongside a traditional state-feedback controller~\cite{pmlr-v155-dean21a}, but in this paper we attempt to solve the more challenging problem where the CBF and controller are defined directly in the space of visual observations. The main issue that arises when defining the CBF as a function of state is poor generalization. Previous works have shown that CBFs defined as functions of state do not generalize well to new environments (e.g. when obstacles have moved), whereas CBFs defined as functions of observations are naturally conditioned on the observed state of the environment~\cite{dawson22perception}.

\section{Problem Statement \& Assumptions}\label{problem_statement}

Our goal in this paper is to adapt the CBF safety filter framework described in Eq~\eqref{cbf_optimization} and~\cite{dawson2022safe,ames_cbf,Agrawal17dt} for use in the visual-feedback setting. Here, we state the control objectives and assumptions used throughout our paper.

As the robot navigates its environment, it builds an implicit scene representation in the form of a NeRF (which is partly pre-trained to enable fast adaptation). As the robot explores, the controller must ensure safety, i.e. avoiding any collisions with the environment. To achieve this goal using a safety filter like Eq.~\eqref{cbf_optimization}, the filter must be able to tell whether an intended control action will bring the robot dangerously close to any nearby obstacle. If the intended action is found to be unsafe, the filter must find the smallest control intervention that still preserves safety. Formally, these requirements are:

\begin{enumerate}
\item \textit{Safety:} the robot must remain a sufficient distance from any obstacles visible nearby; i.e. given an RGBd image $y \in \R^{w \times h \times 4}$, $\min_{0 \leq i \leq w;\ 0\leq j \leq h}y[i, j]_{depth} > d_c$, where $d_c>0$ is the minimum safe clearance.

\item \textit{Minimal intervention:} the safety filter should minimize any deviation from the nominal control policy, insofar as is possible without violating the safety requirement. This requirement is naturally captured by the CBF-QP with objective~\eqref{cbf_optimization} and constraint~\eqref{cbf_constraint}.
\end{enumerate}

We note that our definition of safety is limited to what can be observed by the robot via its cameras, so we do not consider the case when a robot can collide with an obstacle in its blind spot. This is an important case that deserves additional study, but it is outside the scope of this paper; we focus only on the case when safety can be defined purely as a function of the robot's observations. We also note that our approach can be easily extended to the case when the robot has multiple cameras, so covering these blind spots would be as simple as adding additional cameras in practice.

Finally, we would like to design our safety filter so that it can be easily adapted to robots with different dynamics. To do this, we will make use of a \textit{two-level control architecture}, inspired by other works on formal methods for robot safety~\cite{herbert2017fastrack,Chen_Li_Fan_Williams_2021}. In this architecture, we apply our visual-feedback safety filter at the top level with a simplified model of the robot's dynamics (e.g. a single- or double-integrator or a Dubins car), then use a low-level controller to track the top-level state using the full nonlinear dynamics of the system. Several techniques exist for deriving lower-level controllers with guaranteed tracking bounds~\cite{herbert2017fastrack,sun21_c3m}, and so we focus on deriving a CBF visual-feedback safety filter for the high level controller with simplified plant dynamics (the error bound of the tracking controller determines the safety margin $d_c$ in our framework, ensuring that safety is still maintained).

\section{Approach}
\label{approach}

We will first provide an overview of our control architecture, then provide more detail on the two main subsystems: the observation-space CBF used to maintain safety and the NeRF used to predict future values of that CBF.

\subsection{Overview}

\begin{figure}
    \label{fig1}
    \begin{center}
	\includegraphics[width=\linewidth]{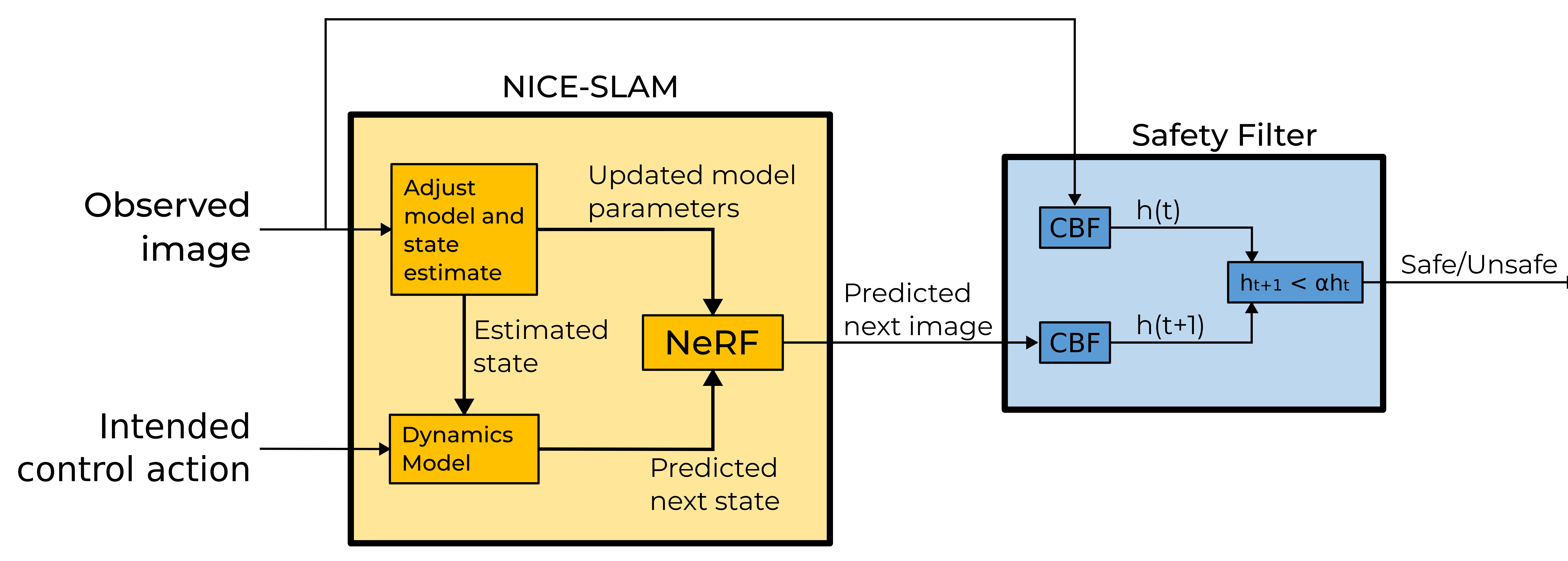}
    \end{center}
    \caption{The NeRF-enabled safety filter used in our control architecture. The NICE-SLAM subsystem~\cite{zhu2022nice} is used to predict future RGBd observations using a neural implicit representation of the scene, and a CBF uses those predictions to decide if a control action is safe. If the intended control action is unsafe, our algorithm searches for nearby control actions that are safe (not shown, but described in Algorithm~\ref{alg:safety_filter}).}
	\label{fig:overview}
\end{figure}

The algorithm for our discrete-time visual CBF safety filter is provided in Algorithm~\ref{alg:safety_filter}, and the interaction between the neural-implicit scene representation (for which we use the NICE-SLAM implementation~\cite{zhu2022nice}) and the CBF is illustrated in Fig.~\ref{fig:overview}. Algorithm~\ref{alg:safety_filter} proceeds as follows: first, the most recent observation is used to update both the state estimate and neural scene representation using the NICE-SLAM subsystem~\cite{zhu2022nice}. Then we predict the next state and observation given the nominal control input $u_t = \pi_0(y_t)$ using the dynamics and neural scene representation, respectively. We then use the CBF to decide whether this control action is safe; if it is, we accept this control action, but if it is not we re-sample randomly in the vicinity of $u_t$ to find new candidate control actions. In practice, instead of re-sampling control actions one-at-a-time if the nominal policy is unsafe, we evaluate multiple candidate interventions in parallel to speed execution time and find a safe candidate close to the nominal control action.

\begin{algorithm}
\caption{The visual-foresight CBF safety filter. $\theta_t$ denotes the model parameters of the neural scene representation, which are fine-tuned at each step.}\label{alg:safety_filter}
    \begin{algorithmic}
        \Require Image $y_t$, nominal policy $\pi_0$, NeRF parameters $\theta_{t-1}$
        \State $u_0, u \gets \pi_0(y_t)$, $\textit{safe} \gets \textsc{False}$
        \State $\theta_{t}, \hat{x}_t \gets \textsc{NSUpdate}(y_t, \theta_{t-1})$ \Comment{update model \& state}
        \While{not \textit{Safe}}
        \State $\hat{x}_{t+1} \gets f(\hat{x}_t, u)$ \Comment{predict state}
        \State $\hat{y}_{t+1} \gets \textsc{NSPredict}(\hat{x}_{t+1}, \theta_t)$ \Comment{predict observation}
        \State $h_t \gets h(y_t),\ h_{t+1} \gets h(\hat{y}_{t+1})$ \Comment{evaluate CBF}
        \If{$h_{t+1} \leq \alpha h_t$}
            \State $\textit{safe} \gets \textsc{True}$, \Return $u, \textit{safe}$
        \Else
            \State $\textit{safe} \gets \textsc{False},\ u \gets \textsc{Re-sample}(u_0)$
        \EndIf
        \EndWhile
    \end{algorithmic}
\end{algorithm}

\subsection{Visual Control Barrier Functions}

The CBF $h(y): \R^{w\times h \times 4} \mapsto \R$ is used to ensure that the system remains safe. In order to be useful, a CBF must not only distinguish safe ($h < 0$) and unsafe ($h > 0$) situations, but it must also generalize well to new environments. This need for generalization is our key motivation for defining the CBF as a function of observations. Previous work~\cite{dawson22perception} has shown that observation-space CBFs generalize much better than state-based CBFs; the intuition is that if an obstacle moves in the scene, then previously-safe states may become unsafe, but an image of the obstacle will usually be a reliable indicator of how far away it is.

In keeping with our use of a two-layer control architecture, as discussed in Section~\ref{problem_statement}, we will define a CBF for two common simplified dynamics models, the single- and double-integrators, which we can then combine with a lower-level tracking controller to ensure the safety of robots with a wide range of possible dynamics~\cite{Chen_Li_Fan_Williams_2021,herbert2017fastrack}. When using first-order dynamics, we use the CBF $h(y) = d_c - \min_{i, j}y[i, j]_{depth}$.
When using second-order dynamics, we can augment the CBF with an estimate of the robot's velocity: $h(y, \hat{v}) = d_c - \min_{i, j}y[i, j]_{depth} + \beta||\hat{v}||$
where $\beta > 0$ is an empirically-tuned hyperparameter.
This CBF allows the robot to extrapolate and intervene to modify the control action even when collision is not imminent (we observe empirically in Section~\ref{result} that the CBF will often gradually increase its intervention as the robot approaches an obstacle).

\subsection{Neural Implicit Scene Representation and State Estimation}

Our safety filter in Algorithm~\ref{alg:safety_filter} relies on the ability to not only estimate the robot's position and build a representation of the scene (\textsc{NSUpdate}) but also predict what observations the robot \textit{could} receive if it were to take a particular action (\textsc{NSPredict}). We rely on recently-developed NeRF-based tools for both of these functions. In particular, we integrate the NICE-SLAM (Neural Implicit Scalable Encoding for SLAM) framework presented in~\cite{zhu2022nice} into our safety filter. The core of NICE-SLAM is a set of pre-trained NeRFs that are retrained online to build a representation of the scene as the robot explores. Since NeRF rendering is differentiable, NICE-SLAM is able to optimize both the parameters of its NeRFs (the mapping step) and an estimate of the robot's pose (tracking); \textsc{NSUpdate} refers to both the mapping and tracking steps.

NICE-SLAM is also able to predict the RGBd image the robot would likely observe if it were to move to a new pose in the environment by simply re-rendering the NeRF from the new viewpoint (\textsc{NSPredict}). This capability allows the robot to imagine what observations it would receive as a result of taking different actions, so the robot can proactively determine whether an action would be unsafe and avoid it. Calls to \textsc{NSPredict} can be easily batched and parallelized on GPU to reduce the time needed to execute Algorithm~\ref{alg:safety_filter}.

\section{Experiments}
\label{exp}

\newcommand{\nerfnav}{\textsc{NeRF-Navigation} }

In this section, we present experimental results to characterize our proposed safety filter, justify its key design decisions, and understand the effect of hyperparameters on controller performance. We compare our controller with the \nerfnav framework, which maintains safety by solving a trajectory optimization problem using the point-wise density of the NeRF scene representation as a proxy for collision cost. To our knowledge, \nerfnav is the only other published work using NeRFs for robot motion control, but we find that it has several performance and reliability issues. Although our work and \nerfnav have different goals (safety filter vs trajectory planning), it is still useful to compare against another NeRF-based approach.

To evaluate our safety filter, we use a simple \textit{teleoperation} setting where a human is remotely piloting the robot through an environment. In this setting, the human provides a convenient source of nominal policy $\pi_0(y)$: the human observes the feed from the robot's camera and makes control decisions. Since these actions may be unsafe, we must use a safety filter to prevent the human from causing a collision. To create a fair comparison, we adapt \nerfnav for this use case by attempting to find a trajectory that stays close to the human's intended motion over a short horizon.

In the following experiments, we are interested in three key metrics: the magnitude of the action intervention $\Delta u = ||u-\pi_0(y)||$, the minimum clearance $d_{min} = \min_{i,j}y[i,j]_{depth}$, and the time required to evaluate the control policy.

\subsection{Experimental Setup}

\paragraph{Environments} We conduct experiments in the \texttt{Room1} indoor environment from the Replica dataset\cite{straub2019replica}.

\paragraph{Test Procedure} We initialize the robot facing towards a wall at a distance of approximately \SI{2.5}{m}. We then provide a worst-case nominal control input driving the robot directly towards the wall (\SI{1}{m/s} for the single integrator and \SI{1}{m/s^2} for the double integrator).

\paragraph{Implementation Details} We conduct our experiments on a desktop PC with a 2200 MHz Intel i7-10700K CPU and an NVIDIA RTX 3090 GPU. For the NICE-SLAM and \nerfnav implementations, we use the hyperparameters provided in their respective source repositories. We run all controllers at a rate of \SI{10}{Hz} for \SI{10}{s}, slowing the simulation clock if necessary (our control method is capable of running in real-time, but \nerfnav is not). We sample candidate actions from a Gaussian distribution $\mathcal{N}(0,I)$, and evaluate in batches of 10. We use parameters $d_c=0.1, \alpha=0.5$ for the single-integrator CBF and $\beta=1$ for the double-integrator CBF. We run all the experiments 5 times and report the average and range for all measurements.

\subsection{Experimental Results}
\label{result}

\begin{figure*}
    \begin{center}
	\includegraphics[width=\linewidth]{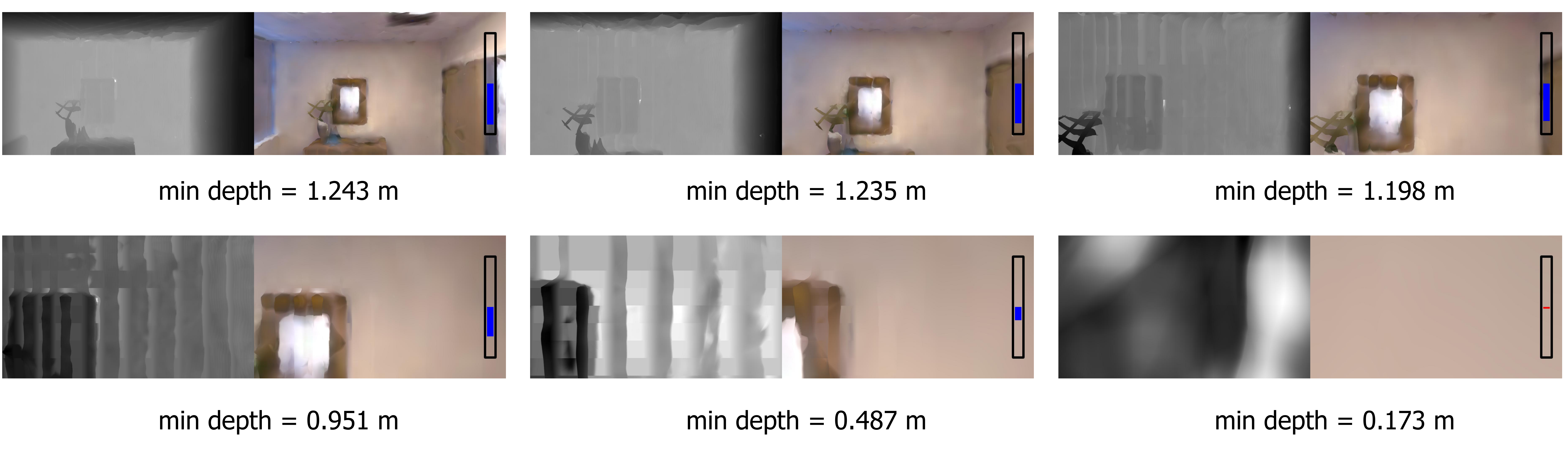}
    \end{center}
    \caption{Depth (left, greyscale) and RGB (right) images rendered by the NeRF showing the effect or our visual CBF safety filter in simulation with abstracted single-integrator dynamics. Images are shown \SI{0.5}{s} apart. The bar on the right side of each image shows the current CBF value (the center of the range is $h = 0$); the bar is blue when the safety filter does not modify the intended action and red when it intervenes. The CBF value decrease as the robot approaches to the wall, and the controller successfully avoids collision. Note that we do not use RGB data for control, only for illustration.}
    \label{fig:result_renders}
\end{figure*}

We study the performance of our CBF safety filter with single- and double-integrator dynamics and compare its performance with that of \nerfnav\cite{Adamkiewicz2022nerf_navigation}.

\begin{figure}
    \centering
    \begin{subfigure}{0.49\linewidth}
    \centering
    \includegraphics[width=\linewidth]{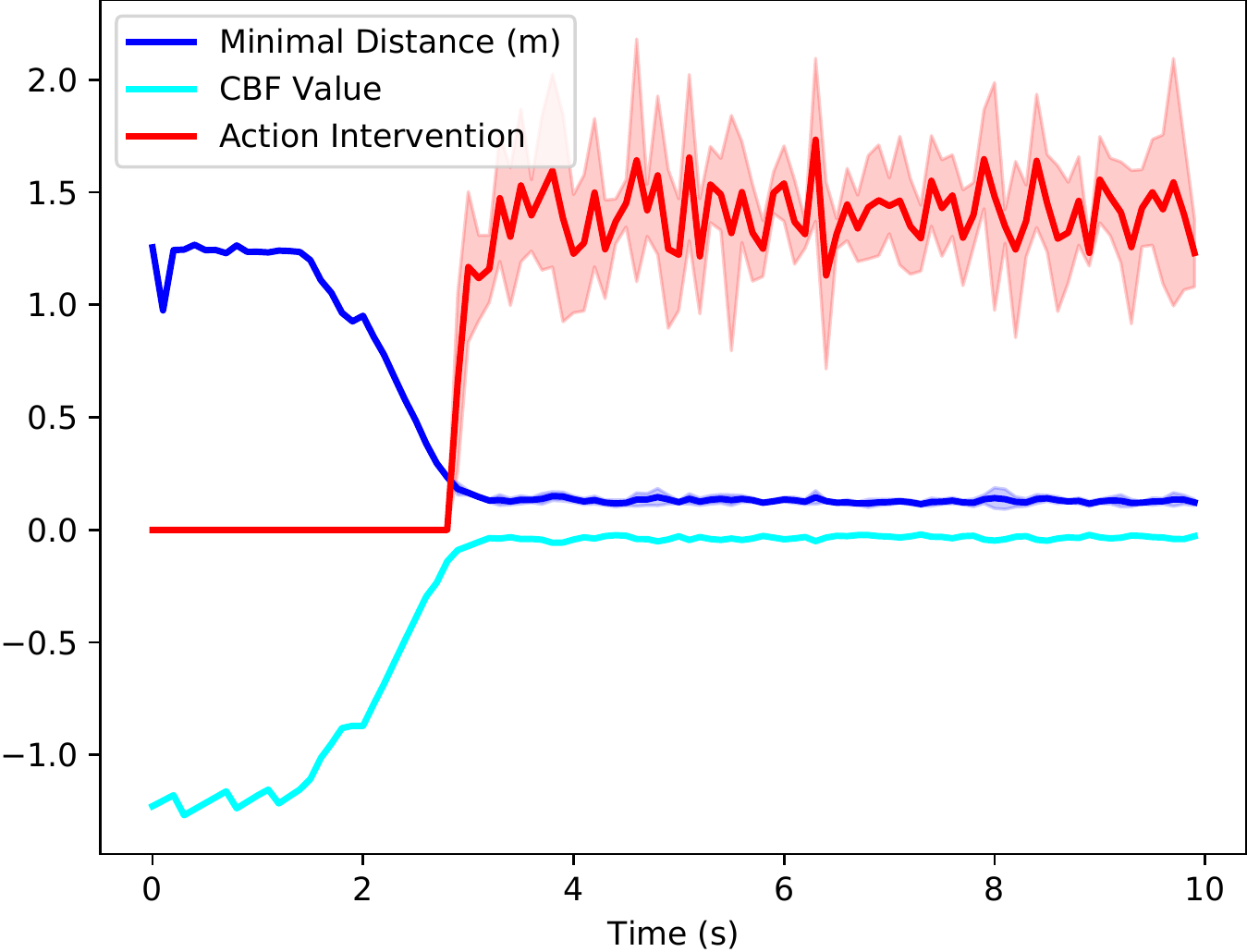}
    \caption{CBF}
    \label{cbf_single_results}
    \end{subfigure}
    \begin{subfigure}{0.49\linewidth}
    \centering
    \includegraphics[width=\linewidth]{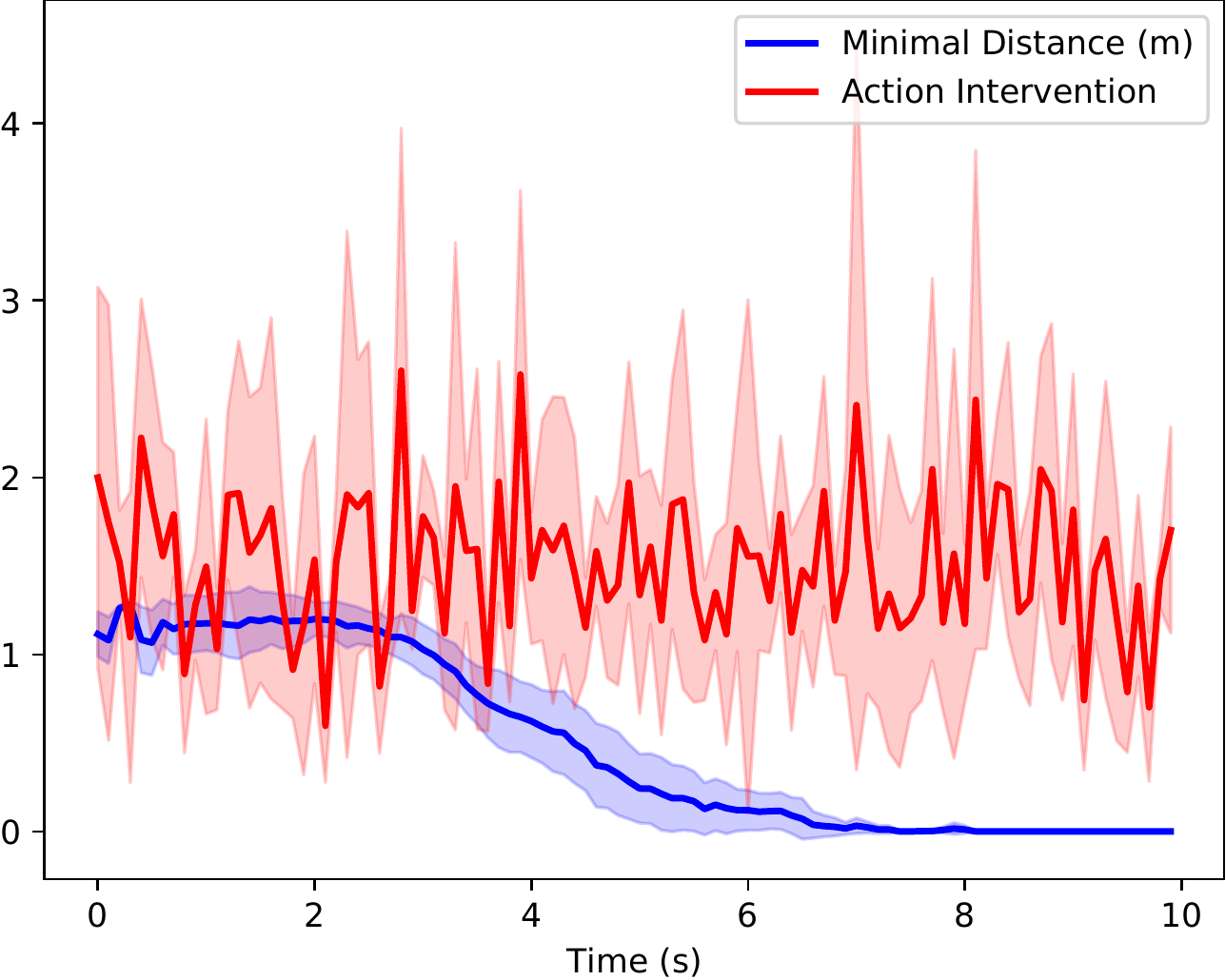}
    \caption{\nerfnav}
    \label{nerf_nav_single_results}
    \end{subfigure}
    \caption{\textbf{(a)} the CBF value and action intervention over time for our CBF controller with single integrator dynamics. \textbf{(b)} a similar plot for \textsc{Nerf-Navigation}. While NeRF-Navigation fails to prevent robot from crashing, our method successfully filters unsafe actions when the robot is close to the wall and avoids unnecessary intervention when it is far away. Shaded areas show standard deviation across 5 runs.}
\end{figure}

\paragraph{Single-integrator System} Fig.~\ref{fig:result_renders} shows a representative series of rendered RGBd images demonstrating how our CBF controller detects when the robot approaches the wall and gradually intervenes to avoid collision. We measure the minimum distance to the nearest obstacle, the CBF value, and the intervention for each time step and plot the results in Fig.~\ref{cbf_single_results}. We see that our CBF safety filter does not intervene until the robot gets within ~0.25 m of the wall, at which point it intervenes to bring the robot to a smooth stop. In contrast, the \nerfnav controller (Fig.~\ref{nerf_nav_single_results}) is much more conservative, applying a non-zero intervention even when the robot is more than 1 m from the wall. Moreover, \nerfnav is not able to prevent the robot from colliding with the wall (0 m minimum separation), which may be due to the fact that the NeRF density (upon which \nerfnav relies) is not a reliable indicator of safety; we explore this point further in Section~\ref{ablation}.

\begin{figure}
    \centering
    \begin{subfigure}[t]{0.49\linewidth}
    \centering
    \includegraphics[width=\linewidth]{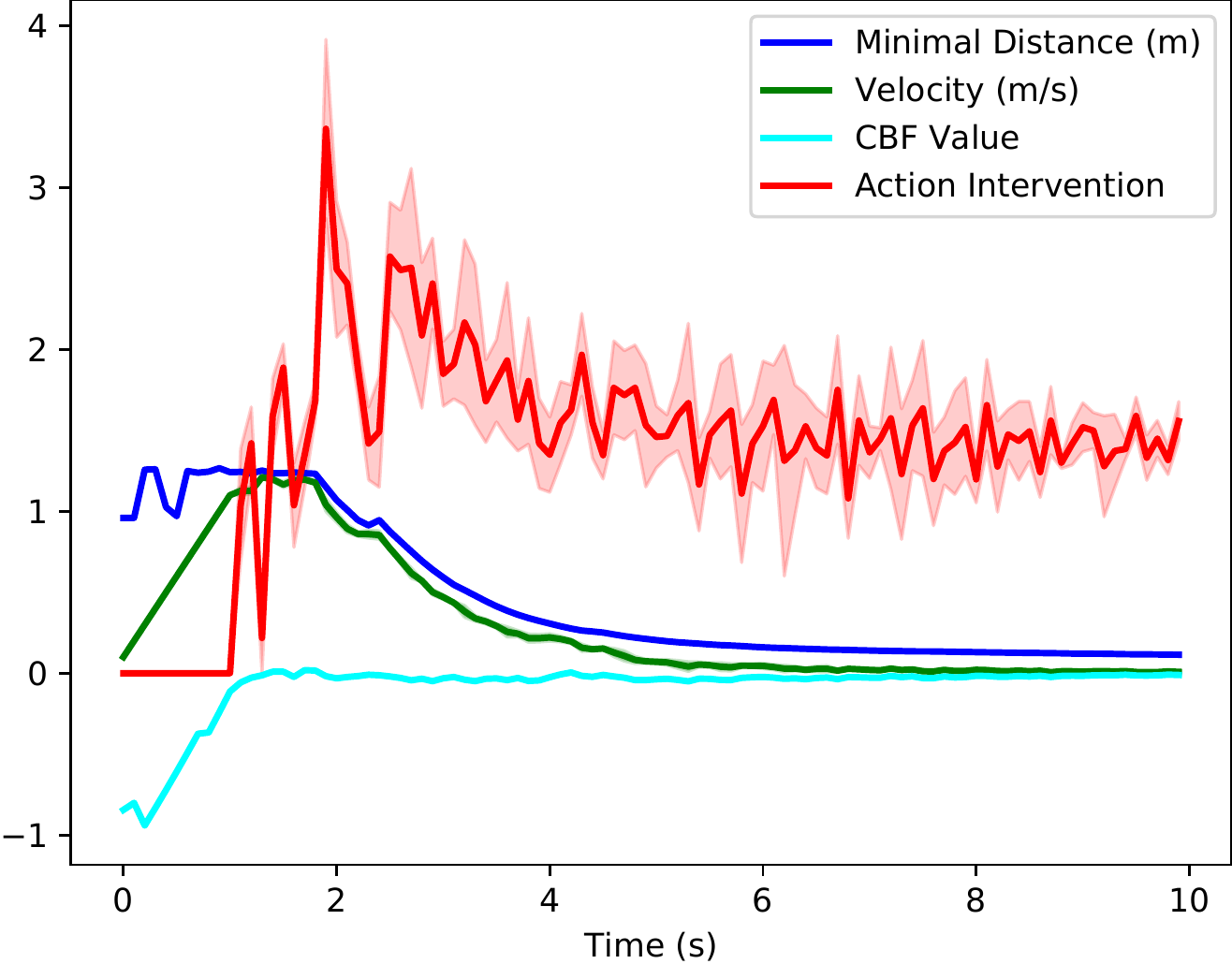}
    \caption{CBF}
    \label{cbf_double_results}
    \end{subfigure}
    \begin{subfigure}[t]{0.49\linewidth}
    \centering
    \includegraphics[width=\linewidth]{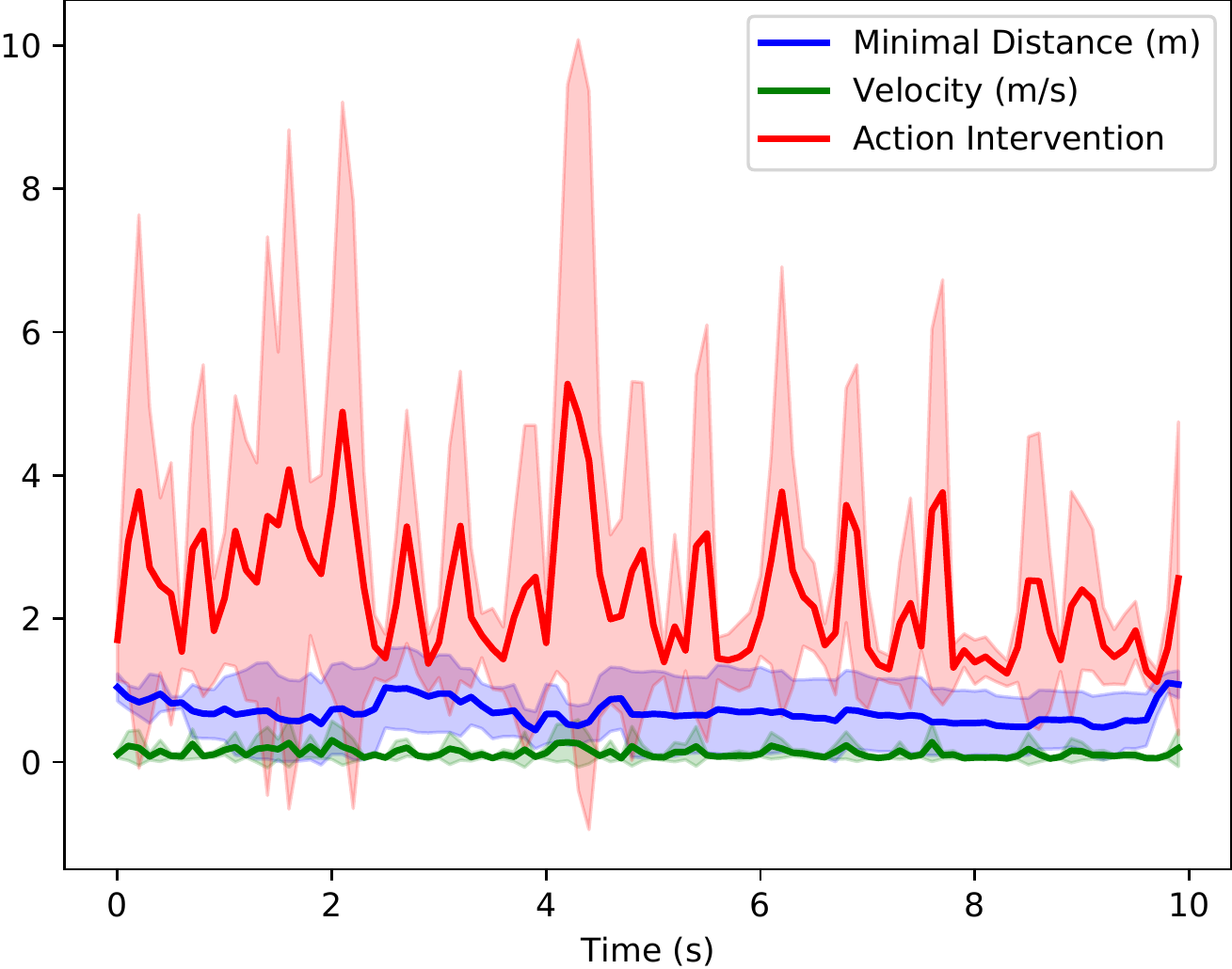}
    \caption{\nerfnav}
    \label{nerf_nav_double_results}
    \end{subfigure}
    \caption{\textbf{(a)} the CBF value, action intervention, velocity, and minimal distance over time for our CBF controller with double integrator dynamics. \textbf{(b)} a similar plot for \textsc{Nerf-Navigation}. \nerfnav is able to stay safe only by applying large interventions and acting conservatively. Our CBF controller is able to gradually increase its intervention as the robot moves faster towards the wall, enabling it to be both safe and not conservative. Shaded areas show standard deviation across 5 runs.}
\end{figure}

\paragraph{Double-integrator System} In these experiments, we observe the velocity of the robot in addition to the quantities discussed above. Fig.~\ref{cbf_double_results} shows the performance of our controller: we see that the controller does not intervene until the CBF value approaches 0, at which point the controller begins to decelerate, bringing the robot to a stop just before the wall. In contrast, the performance of \nerfnav is shown in Fig.~\ref{nerf_nav_double_results}; this controller is much more conservative and does not allow the robot to approach the wall at all.

\paragraph{Runtime} We compare time required to evaluate our controller and \nerfnav at a single step. We found that \nerfnav required more than 80 seconds to find a trajectory, which is not suitable for online use. In contrast, our CBF safety filter can be evaluated in 0.01 s when the nominal control action is safe, and if an intervention is needed then we can evaluate a batch of 10 candidate actions in 0.095 s, suitable for online use at 10 Hz. This significant advantage in runtime stems from our use of CBFs; while \nerfnav must consider a multi-step trajectory optimization problem (ultimately requiring more than a thousand NeRF queries), our CBF-based system uses a one-step horizon and requires only one NeRF query for each candidate action.

\begin{figure}
    \centering
    \begin{subfigure}[tb]{0.49\linewidth}
    \centering
    \includegraphics[width=\linewidth]{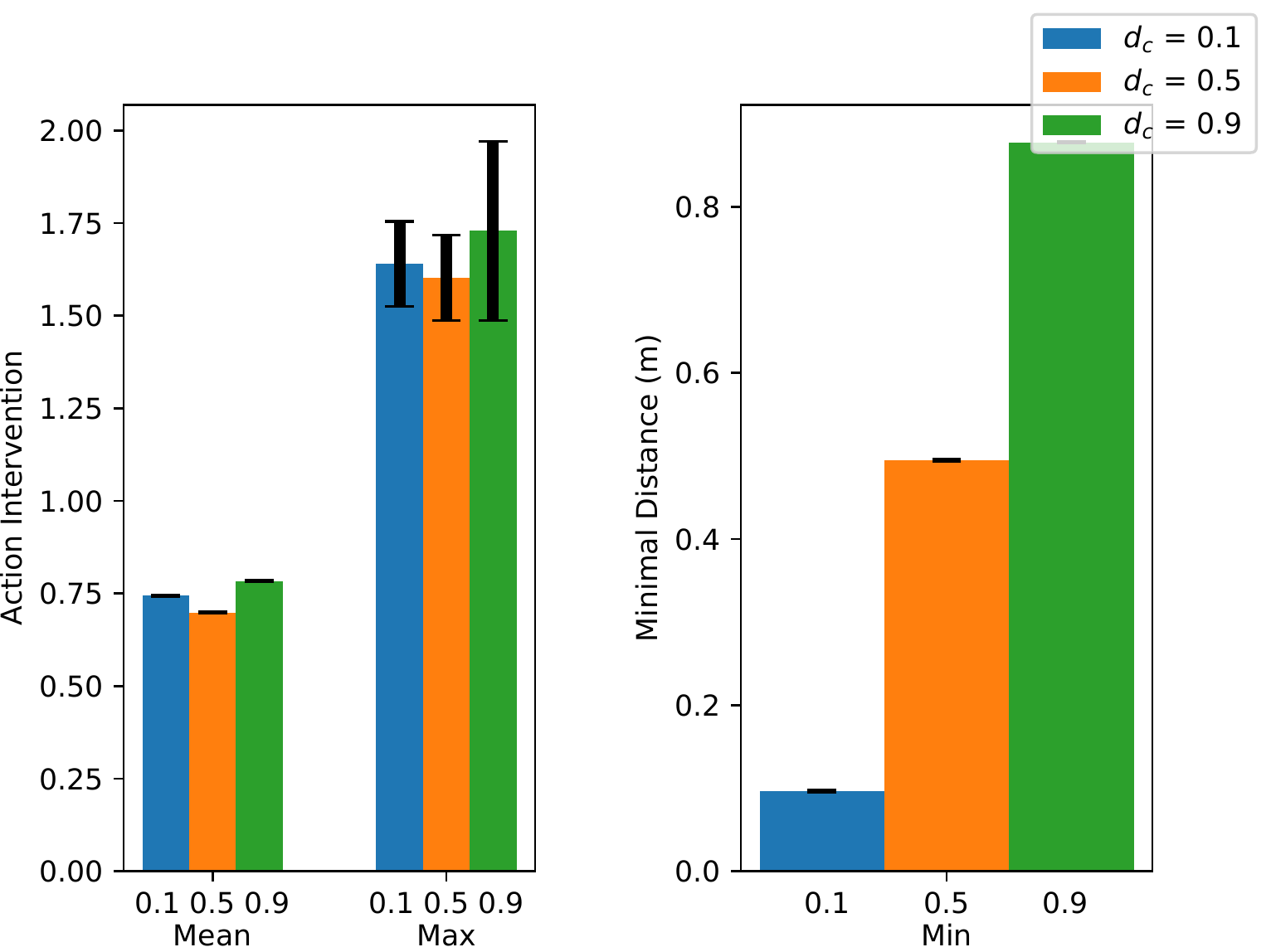}
    \caption{$d_c$ sweep}
    \label{dc_sweep_results}
    \end{subfigure}
    \begin{subfigure}[tb]{0.49\linewidth}
    \centering
    \includegraphics[width=\linewidth]{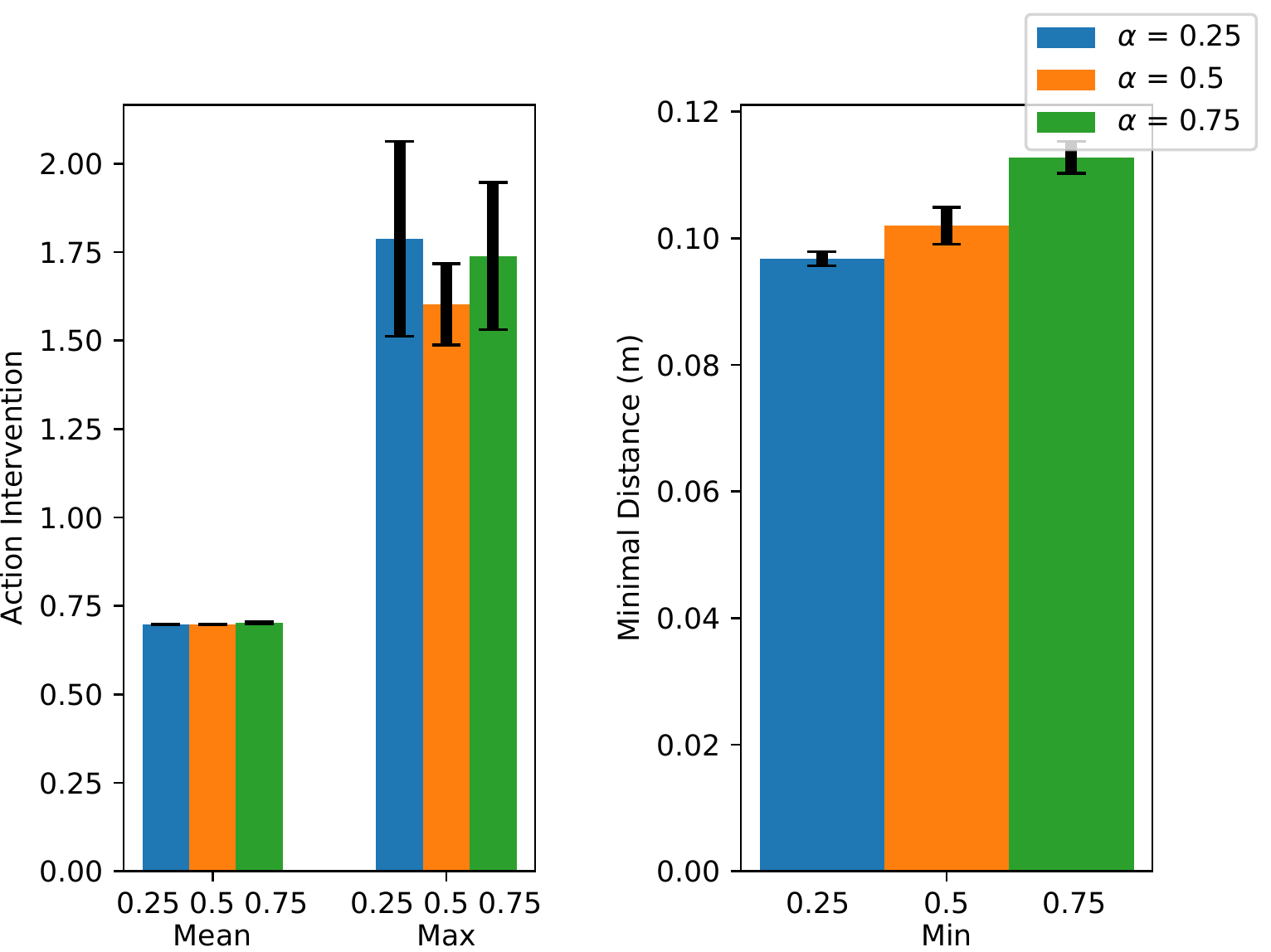}
    \caption{$\alpha$ sweep}
    \label{alpha_sweep_results}
    \end{subfigure}
    \caption{\textbf{(Left)} Mean and maximum intervention and minimum distance to collision for $d_c=\{0.1,0.5,0.9\}, \alpha=0.5$, showing that the minimum distance to collision closely matches $d_c$, validating the effectiveness of our CBF method. \textbf{(Right)} Mean and maximum intervention and minimum distance to collision for $\alpha=\{0.25,0.5,0.75\}, d_c=0.1$. We see that smaller values of $\alpha$ require larger maximum interventions, since the system must slow down sharply at the last minute when $\alpha$ is small. Both plots use single-integrator dynamics.}
\end{figure}

\paragraph{Parameter Sweep} We study the effect of varying $d_c$ (the required safety margin) and $\alpha$ (governing the rate at which the CBF is allowed to approach 0). We test the performance of $d_c=\{0.1, 0.5, 0.9\}$ with constant $\alpha = 0.5$ and $\alpha=\{0.25, 0.5, 0.75\}$ with constant $d_c = 0.5$ and compare the average intervention, maximum intervention, and the minimum distance to the wall observed during the experiment. Fig.~\ref{dc_sweep_results} shows that the minimal distance in each trial matches $d_c$ closely, indicating that the CBF safety filter works as intended.
Fig.~\ref{alpha_sweep_results} shows that decreasing $\alpha$ increases the maximum intervention needed to preserve safety, as the filter only intervenes at the last minute when $\alpha$ is small.

\begin{figure}
    \centering
    \begin{subfigure}[tb]{0.49\linewidth}
    \centering
    \includegraphics[width=\linewidth]{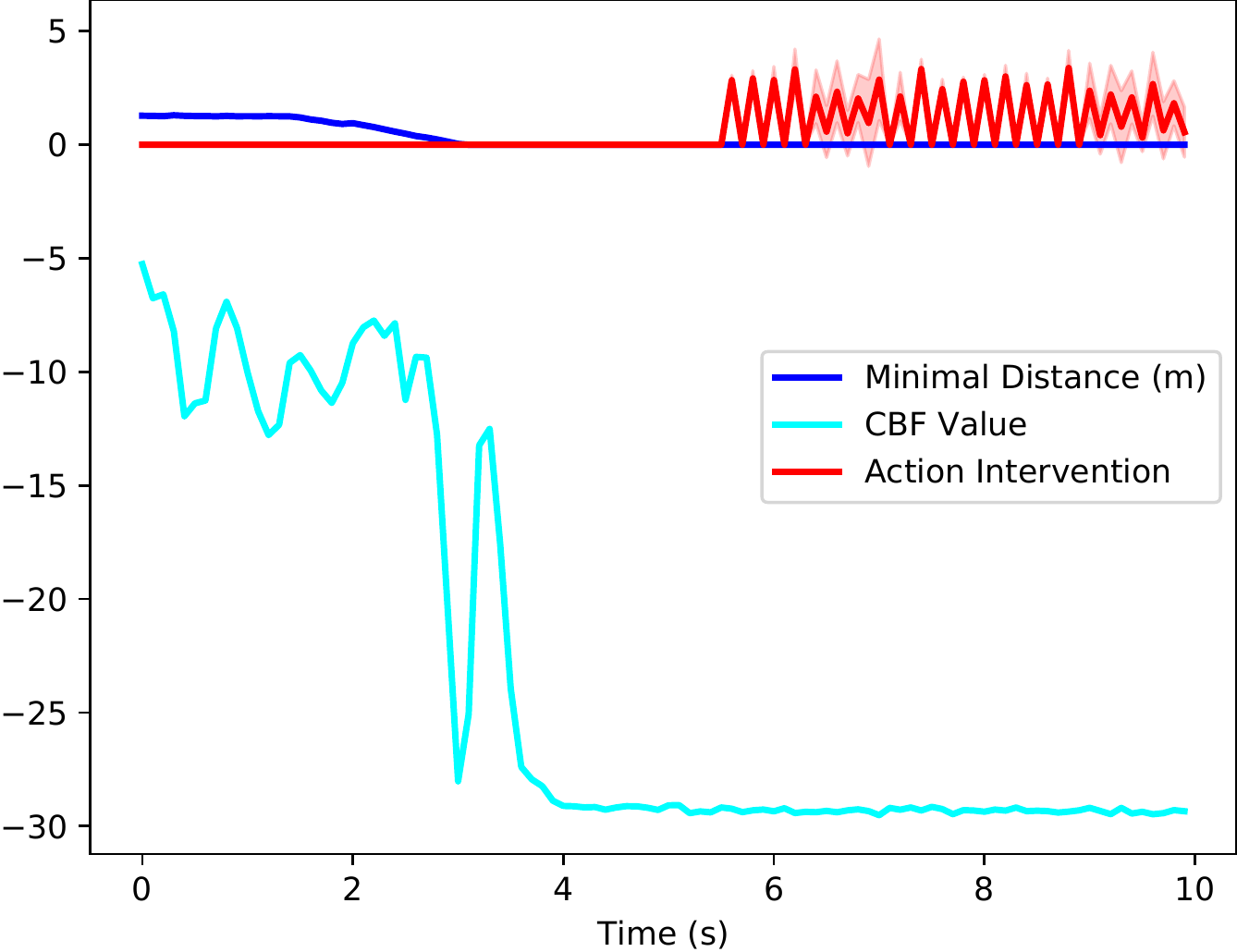}
    \caption{Performance of density CBF}
    \label{density_single_results}
    \end{subfigure}
    \begin{subfigure}[tb]{0.49\linewidth}
    \centering
    \includegraphics[width=\linewidth]{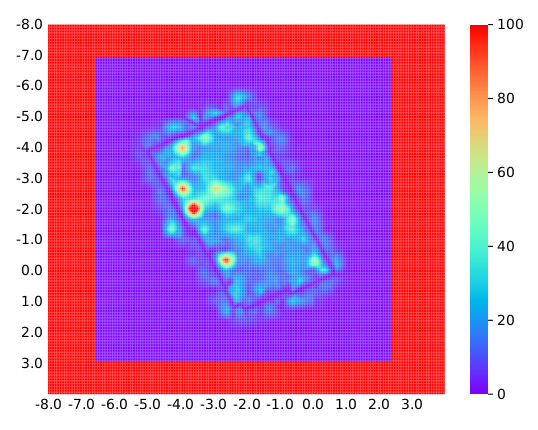}
    \caption{Density map}
    \label{density_map}
    \end{subfigure}
    \caption{\textbf{(Left)} CBF value, intervention, and distance to collision over time for a density-based CBF controller with single integrator dynamics. We see that the CBF remains negative even when the robot collides with the wall, suggesting that NeRF density is not an appropriate indicator of safety. \textbf{(Right)} the density map of the environment at height $z=0.5m$ helps explain this failure: the density map does not accurately capture the boundaries of the walls (density peaks at the edges of the room but then decays at the walls).}
\end{figure}

\subsection{Ablation Study}
\label{ablation}

Our proposed CBF safety filter relies on the rendered RGB and depth images from the NeRF, rather than using the underlying NeRF density field as a proxy for collision likelihood (as \nerfnav does). The NeRF density refers the probability of a point terminating a ray of light, whereas depth is determined by integrating along a ray originating at a camera. In this section we justify our decision to rely on depth by showing that a CBF based on density alone does not preserve safety; we believe this is because the density field is not a consistent proxy for collisions.


To conduct this comparison, we define a density-based CBF $h(\hat{x})=d_c+\sigma(\hat{x})$, where $\sigma(\hat{x})$ is the NeRF density at the location implied by state $x$. Density is positive and tends to increase in occupied regions of space, so we set $d_c$ to a negative value determined by measuring the density near the walls; note that $d_c$ no longer directly refers to a physical safety margin in this setting. Fig.~\ref{density_single_results} shows the performance of a density-based CBF controller with $d_c=-30$; we see that this CBF does not intervene even as the distance between the robot and the wall drops to 0. This behavior appears to be caused by a lack of a clear dependence between NeRF density and distance to the walls. We also test $d_c = -40$ (finding a similar result)  and $d_c = -20$ (unable to find a safe control action by solving Eq.~\eqref{cbf_optimization}). 

To gain a better understanding of the NeRF density field, we plot a 2D slide of the density field at a height $z = 0.5$ m in Fig.~\ref{density_map}. We see that certain obstacles in the environment cause high density, but the NeRF density near the walls is not particularly high. As a result, we conclude that NeRF density is not a reliable indicator of safety.
 
\section{Discussion \& Limitations}

In this paper, we present a novel CBF-based controller that ensures safety for a controller operating solely on high-dimensional visual feedback. We combine CBFs with NeRFs to enable single-step visual foresight, allowing our controller to determine whether an action is safe or unsafe using only a single-step horizon. In comparison with previously-published NeRF-based motion controllers, which cannot run in real-time and can be overly conservative, our controller can run in simulation at real-time rates and maintains safety while limiting the magnitude of its control intervention.

Our work in this paper also presents a number of drawbacks that point to interesting directions of future work. On a practical level, the most crucial point for future work is the acceleration in search of safe actions. Our method can evaluate at most 10 control actions during each step (in order to meet a 10 Hz control frequency requirement). This time is driven largely by the time needed to render depth images from the NeRF; improvements in the ability to optimize NeRFs (even at the expense of resolution) would help our method consider more candidate actions at each step. A second important step for future research is the development of vision-based CBFs for more complex dynamics, for instance by using neural networks to adapt NeRF-based CBFs for new dynamics (as~\cite{dawson22perception} explores in the context of Lidar-based CBFs). Furthermore, although our method can adapt to a limited field of view by adding additional cameras, this would increase the computation time as more images would need to be rendered per step. Consequently, more work is needed to understand how to apply our technique in a limited-computation setting, perhaps by reducing the required image resolution. In future work, we also youhope to explore transferring our method from desktop hardware (where it can run in real time) to common embedded platforms such as the Nvidia Jetson. On a theoretical level, our approach in this paper ignores reconstruction error in the NeRF scene representation, which may negatively impact safety; future work should also explore combining NeRF error bounds (potentially derived from statistical learning theory) with error-robust CBFs~\cite{pmlr-v155-dean21a} to provide more solid safety guarantees.

\bibliographystyle{IEEEtran}
\bibliography{main}

\end{document}